\documentclass[conference]{IEEEtran}
\IEEEoverridecommandlockouts
\usepackage{cite}
\usepackage{amsmath,amssymb,amsfonts}
\usepackage{algorithmic}
\usepackage{graphicx}
\usepackage{textcomp}
\usepackage{xcolor}
\usepackage{booktabs}
\usepackage{multirow}
\usepackage{titlesec}
\def\BibTeX{{\rm B\kern-.05em{\sc i\kern-.025em b}\kern-.08em
    T\kern-.1667em\lower.7ex\hbox{E}\kern-.125emX}}
\begin{document}
\title{FOCUS: Fine-grained Optimization with Semantic Guided Understanding for Pedestrian Attributes Recognition
\vspace{-0.5em}}

\author{\IEEEauthorblockN{Hongyan An$^{1,2,}$\IEEEauthorrefmark{1},
Kuan Zhu$^{2,}$\IEEEauthorrefmark{1},
Xin He$^{1,2}$, 
Haiyun Guo$^{1,2,}$\IEEEauthorrefmark{2}, 
Chaoyang Zhao$^{2,5}$, 
Ming Tang$^{2}$,
Jinqiao Wang$^{1,2,3,4,5}$\IEEEauthorrefmark{2}}
\IEEEauthorblockA{$^{1}$School of Artificial Intelligence, University of Chinese Academy of Sciences}
\IEEEauthorblockA{$^{2}$Foundation Model Research Center, Institute of Automation, Chinese Academy of Sciences}
\IEEEauthorblockA{$^{3}$Peng Cheng Laboratory, $^{4}$Wuhan AI Research, $^{5}$Objecteye Inc.}

anhongyan2022@ia.ac.cn, \{kuan.zhu, haiyuan.guo, jqwang\}@nlpr.ia.ac.cn
\thanks{\IEEEauthorrefmark{1}Equal contribution. \IEEEauthorrefmark{2}Corresponding author.

This work is supported by Beijing Natural Science Foundation under Grant 4244099, Postdoctoral Fellowship Program of CPSF under Grant GZC20232996, China Postdoctoral Science Foundation under Grant 2024M753498, National Natural Science Foundation of China under Grant 62276260, 62176254, Aeronautical Science Foundation of China under Grant 2024M0710M0002.
}
\vspace{-0.5em}
}
\maketitle
\begin{abstract}
Pedestrian attribute recognition (PAR) is a fundamental perception task in intelligent transportation and security. To tackle this fine-grained task, most existing methods focus on extracting regional features to enrich attribute information. However, a regional feature is typically used to predict a fixed set of pre-defined attributes in these methods, which limits the performance and practicality in two aspects: 1) Regional features may compromise fine-grained patterns unique to certain attributes in favor of capturing common characteristics shared across attributes. 2) Regional features cannot generalize to predict unseen attributes in the test time. In this paper, we propose the \textbf{F}ine-grained \textbf{O}ptimization with semanti\textbf{C} g\textbf{U}ided under\textbf{S}tanding (FOCUS) approach for PAR, which adaptively extracts fine-grained attribute-level features for each attribute individually, regardless of whether the attributes are seen or not during training. Specifically, we propose the Multi-Granularity Mix Tokens (MGMT) to capture latent features at varying levels of visual granularity, thereby enriching the diversity of the extracted information. Next, we introduce the Attribute-guided Visual Feature Extraction (AVFE) module, which leverages textual attributes as queries to retrieve their corresponding visual attribute features from the Mix Tokens using a cross-attention mechanism. To ensure that textual attributes focus on the appropriate Mix Tokens, we further incorporate a Region-Aware Contrastive Learning (RACL) method, encouraging attributes within the same region to share consistent attention maps. Extensive experiments on PA100K, PETA, and RAPv1 datasets demonstrate the effectiveness and strong generalization ability of our method. 
\end{abstract}


\section{Introduction}
\label{sec:intro}

Pedestrian attribute recognition (PAR) is a crucial task in the field of human-centric perception\cite{percep,sscr,mgm}, focusing on transforming pedestrian characteristics into a structured representation of various attributes, such as gender, age, clothing style, etc. By identifying these attributes, PAR plays a pivotal role in intelligent transportation applications, such as pedestrian tracking\cite{pedestarin_search}, behavior analysis\cite{PGDM}, and traffic violation identification. It serves as a foundational task that enhances the understanding of complex traffic environments, improving the accuracy of decision-making and the safety of pedestrians.

\begin{figure}[tbp]
\centering
\includegraphics[width=0.45\textwidth]{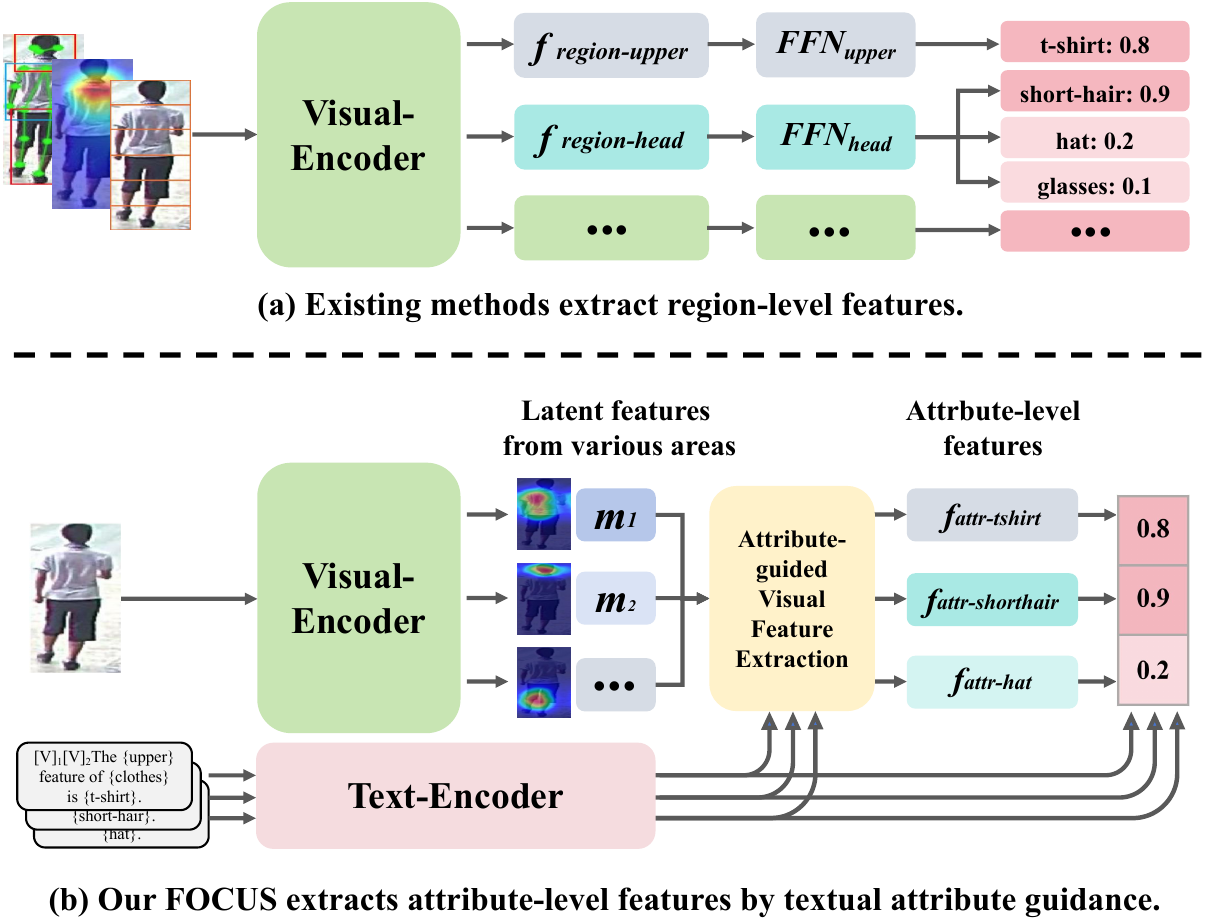} 
\caption{The illustration of feature extraction pipelines in existing methods and our FOCUS. (a) Existing methods extract region-level features through image partitioning, human pose estimation, or attention mechanisms, and train multi-class classifiers to predict a fixed set of pre-defined attributes. (b) Our FOCUS adaptively extracts the fine-grained attribute-level feature for each attribute individually, even if the attribute is not seen during training. }
\label{work_compare}
\end{figure}

The challenges of PAR mainly lie in the diverse and subtle variations in pedestrian appearance, which require the model to extract discriminative and fine-grained features. Existing methods mainly focus on extracting region-specific features to provide more fine-grained information and subsequently predicting the corresponding attributes, based on the assumption that each attribute is typically associated with a particular image region. 
As shown in Fig. \ref{work_compare}(a), existing methods usually use the horizontal stripe-based image partitioning \cite{POAR,parclip}, auxiliary detection modules \cite{PGDM,lgnet_roi}, or attention mechanisms \cite{sscr,attn4} to locate regional areas and extract regional features. 
However, a regional feature is typically used to predict a fixed set of pre-defined attributes in these methods, which limits the performance and practicality in two aspects. On the one hand, regional features may compromise fine-grained patterns unique to certain attributes in favor of capturing common characteristics shared across attributes. For example, the `short-hair', `hat', and `glasses' attributes are all predicted using the regional feature of the head area. To recognize these three unrelated attributes, the head region’s feature may compromise and lose fine-grained information unique to each attribute, potentially leading to inaccurate attribute recognition.
On the other hand, these methods train a multi-class classifier with fixed prediction classes, which prevents them from generalizing to unseen attributes during testing, limiting their practicality.

In this paper, we propose the \textbf{F}ine-grained \textbf{O}ptimization with semanti\textbf{C} g\textbf{U}ided under\textbf{S}tanding (FOCUS) approach for PAR, which can adaptively extract fine-grained attribute-level features for each attribute individually, regardless of whether the attributes are seen or not during training. 
Specifically, we first introduce the Multi-Granularity Mix Tokens (MGMT) to extract diverse features from pedestrian images, creating a latent feature space for subsequent attribute-level feature extraction. The Mix Tokens are additional learnable parameters similar to the Class Token, but each Mix Token interacts with distinct area of input images, enabling the model to extract fine-grained features from various regions. 
Next, we propose the Attribute-guided Visual Feature Extraction (AVFE) module to extract attribute-level features from the Mix Tokens, with each attribute-level feature directly used to predict its corresponding attribute. The textual attributes, augmented with learnable prompts, are treated as queries to retrieve relevant visual information from the Mix Tokens via the cross-attention mechanism. The resulting features are considered as attribute-level representations for the specific attributes. Subsequently, the recognition result for each attribute is obtained by calculating the similarity between its textual attribute feature and the corresponding visual attribute-level feature. It is noted that our method supports the input of unseen attributes during training and extracts the corresponding attribute features for recognition.
Additionally, we propose the Region-Aware Contrastive Learning (RACL) to ensure the attributes focus on the correct Mix Tokens, further refining attribute-level features. We apply contrastive learning to the attention maps in AVFE, based on the reasonable assumption that the attributes within the same region should retrieve information from similar Mix Tokens, i.e., attention maps for attributes within the same region are encouraged to align, while those from different regions are expected to diverge. By this way, if an attribute focuses on the wrong Mix Tokens, it can be corrected by other attributes within the same region.
Finally, FOCUS can adaptively extract fine-grained attribute-related information for textual attributes, even if the attributes are not seen during training.

To summarize, the contributions of this paper are as follows: 
\begin{itemize}
  \item We propose the \textbf{F}ine-grained \textbf{O}ptimization with semanti\textbf{C} g\textbf{U}ided under\textbf{S}tanding (FOCUS) approach for PAR, which adaptively extracts the fine-grained attribute-level feature for each attribute individually to recognize it, regardless of whether the attribute is seen or not during training.
  \item We introduce the MGMT and AVFE modules to extract attribute-relevant information from diverse latent features by the guidance of textual attributes. Additionally, a novel loss called RACL ensures that attributes focus on the correct Mix Tokens through contrastive learning, further refining attribute-level features.
  \item Extensive experiments demonstrate the effectiveness of FOCUS, which achieves state-of-the-art performance in both closed and open scenarios on three PAR datasets, i.e., PA100K, PETA, and RAPv1.
\end{itemize}

\section{Related Work}
\subsection{Pedestrian Attribute Recognition}
The existing methods for fine-grained attribute feature extraction in PAR can be divided into part-based \cite{POAR,parclip, PGDM,lgnet_roi} and attention-based\cite{attn4, sofa, fangx} approaches, similar to most pedestrian tasks\cite{lscs,WBC,pass}. For instance, PGDM \cite{PGDM} employs a pre-trained human pose estimator to localize body parts. Similarly, LG-Net\cite{lgnet_roi} utilizes a region detection module to identify attribute-related regions.  Additionally,\cite{attn4} proposes three distinct attention mechanisms—parsing, label, and spatial attention—to capture relevant features. SSCR\cite{sscr} introduces a Spatial and Semantic Consistency framework, utilizing complementary regularizations to capture spatial and semantic relationships across images. These methods primarily extract region-level features for predefined attributes. In contrast, our approach enables more fine-grained, attribute-level feature extraction for open-domain attribute recognition.
\subsection{Vision-Language Learning}
Vision-language pre-training (VLP) has significantly improved the performance of many downstream tasks by aligning image representations with text embeddings in a shared space. Large-scale vision-language pre-training models, such as CLIP \cite{clip}, are trained on vast amounts of image-text pairs by contrastive learning. This pre-training empowers these models with strong open-vocabulary classification capabilities. Furthermore, CoOp\cite{coop} introduces learnable prompt optimization, leveraging prompt-based learning to fine-tune models for specific tasks, demonstrating potential in vision-language applications. In the field of PAR, VTB\cite{VTB} was the first to employ independent vision and text encoder to extract and fuse multimodal features for attribute prediction. PromptPAR\cite{parclip} enhances the multimodal features by prompt learning based on CLIP\cite{clip}. Unlike these methods, our approach utilizes textual attribute guidance to enable the model to adaptively select attribute-relevant information, achieving attribute recognition in open-domain scenarios. The most related method is POAR\cite{POAR}, but the Masking the Irrelevant Patches method neglects attribute information outside the fixed regions. In addition, our approach not only considers multi-granularity information but also achieves attribute-level feature extraction.

\begin{figure*}[tbp]
\centering
\includegraphics[width=0.9\textwidth]{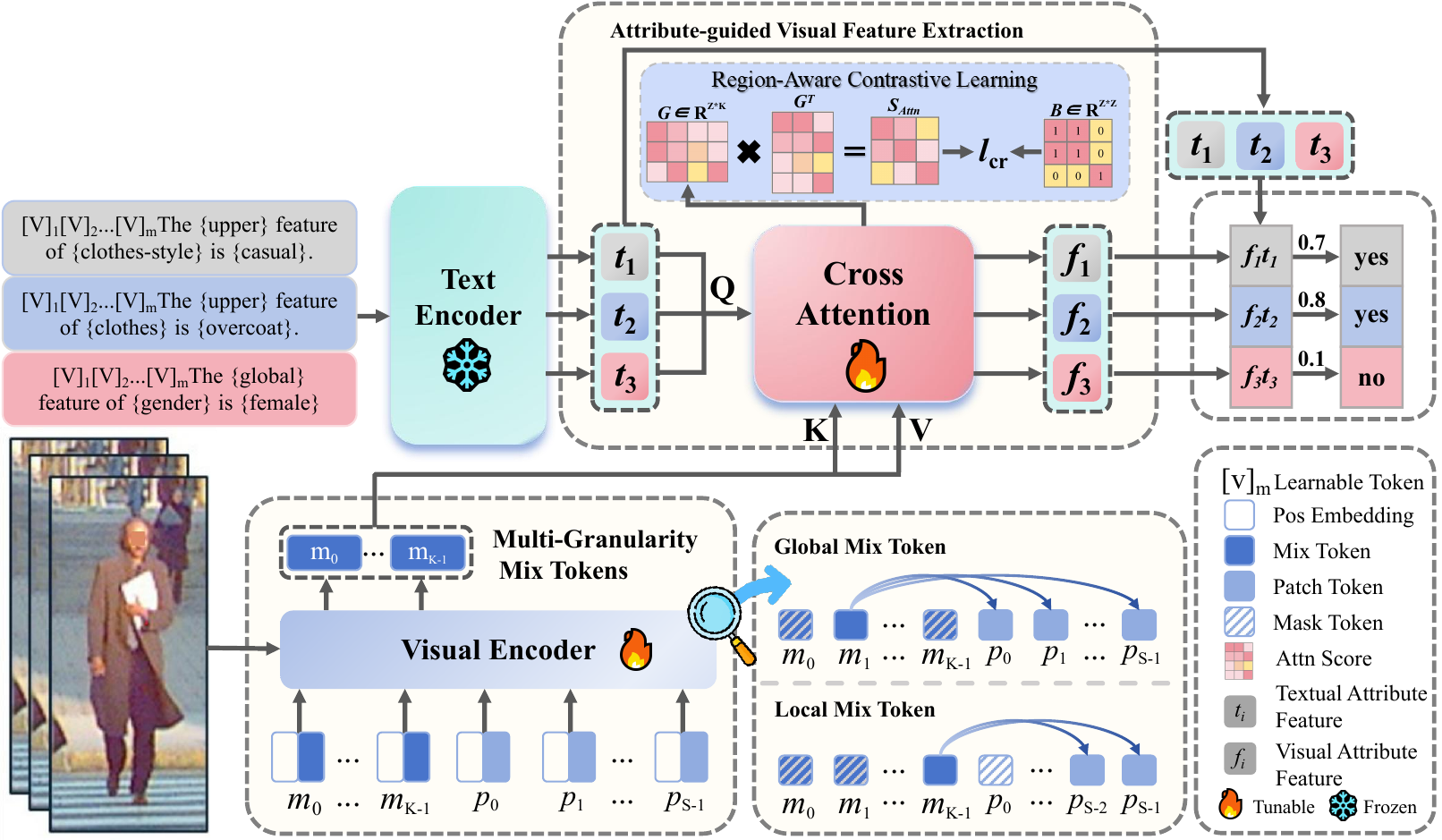} 
\caption{The overview of FOCUS. The Multi-Granularity Mix Tokens module learns diverse features by Mix Tokens interacting with different patch tokens. Then, the textual attributes with learnable prompts are treated as queries to extract the attribute-level features from latent diverse features by Attribute-guided Visual Feature Extraction module, and the Region-Aware Contrastive Learning loss refines the attribute-level features by focusing on the correct Mix Tokens.}
\label{fig_method}
\vspace{-1.5em}
\end{figure*}

\section{Methods}

\subsection{Preliminaries and Model Overview}

The predefined attribute set is denoted as $\mathcal{A} = \{A_1, A_2,...,A_Z\}$, where Z is the total number of attributes and $A_z$ represents the $z^{th}$ specific attribute. A PAR dataset which contains N pedestrian samples is denoted as $\mathcal{D} = \left\{\left(X_{i}, Y_{i}\right)\right\}_{i=1}^{N}$, where $X_{i} \in \mathbb{R}^{W\times H\times 3}$ and $Y_{i} \in\{0,1\}^{Z}$ denote the $i^{th}$ pedestrian image and its attribute label, respectively. The main objective of PAR is to train a model that can identify which attributes of the predefined set $\mathcal{A}$ appear in the given image X.

The CLIP\cite{clip} model provides a feasible approach for achieving open-domain attribute recognition. The visual encoder $\mathcal{V}(\cdot)$ takes a pedestrian image X as input and output the visual feature $\mathcal{V}(X)$. The text encoder $\mathcal{T}(\cdot)$ takes a tokenized attribute description $T_j$ as input, where $T_j$ is obtained through embedding the attribute $j \in \mathcal{S}$ ($\mathcal{S} \supseteq \mathcal{A}$) into a hand-crafted prompt, and outputs the textual attribute feature. The similarity of attribute j and image X can be represented as:
\begin{equation}
\text{Sim}(X,j) = \mathcal{V}(X) \cdot \mathcal{T}^T(T_j)
\end{equation}

A higher similarity indicates a greater probability that the person possesses the corresponding attribute. Compared with CLIP\cite{clip}, we guide the model extract more fine-grained, attribute-level features by textual attributes. As shown in Fig. \ref{fig_method}, the overall framework of FOCUS consists of two components, i.e., MGMT Module and AVFE Module. In the MGMT module, the Mix Tokens capture diverse features by interacting with global image or distinct local image areas. Then, the textual attributes are treated as queries to retrieve the attribute-related visual information from the output Mix Tokens in AVFE and the RACL loss is designed to enhance the correctness of this searching process.

\subsection{Multi-Granularity Mix Tokens}
To provide rich and detailed information for PAR task, we introduce learnable Mix Tokens $\mathcal{M} = [m_0, m_1, ..., m_{K-1}]\in\mathbb{R}^{K\times D}$ to extract fine-grained and diverse features from pedestrian images, where K denotes the number of Mix Tokens and D is the embedding dimension. Similar to the CLS token, the Mix Tokens are learnable parameters and learn to be the visual representations by interacting with patch tokens $\mathcal{P} = [p_0, p_1, p_{S-1}]\in\mathbb{R}^{S\times D}$, where S denotes the number of patch tokens, in self-attention layers. To ensure the learned visual representations diverse, We evenly partition the patch tokens $\mathcal{P}$ into r subsets $\Tilde{\mathcal{P}}$ = [$\Tilde{p}_1$, $\Tilde{p}_2$, ...,$\Tilde{p}_{r-1}$]. Then, we design two types of Mix Tokens: (1) Global-level Mix Tokens $\mathcal{M}_g$, which interact with all patch tokens in the image through self-attention to capture global information. (2) Local-level Mix Tokens $\mathcal{M}_l$, which interact with different subsets of patch tokens $\Tilde{\mathcal{P}}$ to learn different and fine-grained information. As shown in Fig. \ref{fig_method}, the learning process of the two types of Mix Tokens in self-attention can be represented as follows:
\begin{equation}
    \text{Attn}_{g}(\!\mathcal{M}_g,\! \Tilde{\mathcal{P}})\!\! =\!\! \text{Softmax}\!\!\left(\!\!\frac{\mathcal{M}_g \Tilde{\mathcal{P}}^T_K}{\sqrt{d}}\!\right)
    \!\!\Tilde{\mathcal{P}}_V , \Tilde{\mathcal{P}}\!\! =\!\! [\Tilde{p}_1, \Tilde{p}_2, ...,\Tilde{p}_{r_1}]
    \label{m_attention}
\end{equation}
\begin{equation}
    \text{Attn}_{l}(m^l_i, \Tilde{p}_i)\!\! =\!\! \text{Softmax}\!\!\left( \frac{m^l_i (\Tilde{p}_i)^T_K}{\sqrt{d}} \right)\!\! (\Tilde{p}_i)_V, m^l_i\!\!\in\!\! \mathcal{M}_l , \Tilde{p}_i\!\! \in\!\! \Tilde{\mathcal{P}}
\end{equation}
\noindent where $\Tilde{\mathcal{P}}_K$, $\Tilde{\mathcal{P}}_V$ represent the key and value mappings for patch tokens $\Tilde{\mathcal{P}}$, and the same applies to $\Tilde{p}_i$.

To further enhance the diversity of features learned by the Mix Tokens, we introduce a contrastive learning mechanism. Specifically, we calculate the similarity $\mathcal{S}_{mix}$ between different Mix Tokens. Then, we employ a unit matrix $\mathcal{I}\in\mathbb{R}^{K\times K}$ and impose a constraint by calculating the binary cross-entropy loss between different output Mix Tokens and constraint them to be dissimilar. The unit matrix encourages the Mix Tokens to capture distinct information of the image, ensuring that the learned features of different Mix Tokens to be complementary rather than redundant. Formally, the above constraint can be expressed as:
\begin{align}
        \mathcal{L}_{sim} = -\frac{1}{K^2} \sum_{i=1}^{K} \sum_{j=1}^{K} I_{ij} \log(S_{mix(ij)}) \notag \\
    + (1 - I_{ij}) \log(1 - S_{mix(ij)}) \label{eq:sim_loss}
\end{align}

\subsection{Attribute-guided Visual Feature Extraction}

To enable the model to adaptively extract attribute-level features through textual attributes whether seen or unseen, we propose the Attribute-guided Visual Feature Extraction(AVFE) module, which primarily consists of an attribute-guided cross-attention mechanism. 

We expect the textual attributes to more effectively guide attribute-level feature extraction. To this end, we introduce m learnable prompts aligned with each attribute category to enhance the discriminability of semantic information between different attributes. Additionally, we expand the attribute phrase into a textual description with region-specific information to ensure more precise prompts for attribute information. e.g., the final textual description of `T-shirt' is designed as `$[V]_1[V]_2...[V]_m$ a \underline{upper} feature of \underline{clothes} is \underline{T-shirt}.'.  

In the cross-attention operation, the output $\mathcal{T}(t_j)$ of the text encoder is treated as \textit{queries}, and the multi-granularity Mix Tokens $\mathcal{M}_{out}$ extracted by MGMT are regarded as \textit{keys} and \textit{values}. Through this cross-attention mechanism, each attribute query selectively extracts relevant visual information from the Mix Tokens, generating visual features that are specific to that attribute. These attribute-specific visual features are then utilized to determine whether the image possesses the corresponding attribute. Consequently, we designate the extracted features as attribute-level features. The visual attribute-level feature $\mathcal{V}_{t_j}$ and attention maps $G_{t_j}$ of textual attribute $t_j$ can be formulated below:
\begin{equation}
    \mathcal{V}_{t_j}, G_{t_j} = \text{Softmax}\left( \frac{\mathcal{T}(t_j) (\mathcal{M}_{out})^T_K}{\sqrt{D}} \right) (\mathcal{M}_{out})_V \label{cr_attention}
\end{equation}

Although the cross-attention mechanism enables each textual attribute to extract relevant information, we observed that different attributes may not sufficiently concentrate on the most pertinent Mix Tokens, leading to suboptimal feature extraction. To address this issue, we introduce a novel loss function, Region-Aware Contrastive Learning (RACL) loss, with the reasonable assumption that the visual attribute-level features within the same region should focus on identical Mix Tokens, whereas attributes from distinct regions should focus on different Mix Tokens. As shown in Fig. \ref{fig_method}, RACL calculates the similarity between the attention maps G in the cross-attention operation, which is denoted as $\mathcal{S}_{Attn}$. Then, we employ a block matrix $B \in\mathbb{R}^{Z\times Z}$, which enforces higher similarity for textual attributes within the same region and lower similarity for attributes from different regions. The block matrix can be defined as follows:
\begin{equation}
B_{ij} =
\begin{cases} 
1, & \text{if attribute } i \text{ and } j \text{ are in the same region} \\ 
0, & \text{otherwise}
\end{cases}
\end{equation}

By minimizing the binary cross-entropy loss between the similarity matrix and the block matrix, RACL indirectly enhances the ability of textual attributes to focus on the correct Mix Tokens. If the query for a specific attribute incorrectly focuses on the wrong tokens, the attention can be corrected by leveraging the focus of other attributes within the same region, ensuring more accurate alignment. This targeted attention improves the discriminative power of visual attribute-level features and enhances the model's robustness in open-domain scenarios. The RACL is formulated as:
\begin{align}
    \mathcal{L}_{\text{racl}} = - \frac{1}{Z^2} \sum_{i=1}^Z \sum_{j=1}^Z \big[ B_{ij} \log(S_{\text{Attn}(ij)}) \notag \\
    + (1 - B_{ij}) \log(1 - S_{\text{Attn}(ij)}) \big] \label{eq:racl}
\end{align}

\subsection{Loss Function}
Following the loss function of POAR\cite{POAR}, we also use the Many-to-Many Contrastive Loss in the final stage of the training, which consists of two main components: a visual-to-text contrastive branch $\mathcal{L}_{v2t}$ and a text-to-visual contrastive branch $\mathcal{L}_{t2v}$. The final loss function is defined as:
\begin{equation}
\mathcal{L}=\mathcal{L}_{sim}+\mathcal{L}_{racl}+\mathcal{L}_{v2t}+\mathcal{L}_{t2v}.
\label{loss:all}
\end{equation}

We perform attribute prediction by calculating the similarity between textual attribute features and their corresponding visual attribute-level features. As a result, the model is capable of recognizing and associating pedestrian attributes effectively, even in open-domain scenarios where the attributes might not have been seen during training.

\begin{table*}[!htb]
\renewcommand\arraystretch{1.2}
\center
\caption{Comparison with SOTA methods on PETA, PA100K and RAPv1 datasets. Methods in the 1st group are the classifier-based methods. Methods in the 2nd group are the clip-based methods. The first and second highest scores are represented by \textbf{bold} font and `\underline{\hphantom{A}}', respectively. ` $^*$ ' means the re-implementation of this paper with the officially released codes.} 
\vspace{-1.0em}
\label{tab:compare} 
\resizebox{1\textwidth}{!}{
\begin{tabular}{c|c|ccccc|ccccc|ccccc}
\hline \toprule [0.5 pt] 
\multicolumn{1}{c|}{\multirow{2}{*}{Methods}} & \multicolumn{1}{c|}{\multirow{2}{*}{Publish}}   & \multicolumn{5}{c|}{PETA} & \multicolumn{5}{c|}{PA100K} & \multicolumn{5}{c}{RAPv1}\\ \cline{3-17} 
\multicolumn{1}{c|}{} &
\multicolumn{1}{c|}{} &
  \multicolumn{1}{c}{mA} &
  \multicolumn{1}{c}{Acc} & 
  \multicolumn{1}{c}{Prec} &
  \multicolumn{1}{c}{Recall} &
  \multicolumn{1}{c|}{F1} &
  \multicolumn{1}{c}{mA} &
  \multicolumn{1}{c}{Acc} &
  \multicolumn{1}{c}{Prec} &
  \multicolumn{1}{c}{Recall} &
  \multicolumn{1}{c|}{F1}  &
  \multicolumn{1}{c}{mA} &
  \multicolumn{1}{c}{Acc} &
  \multicolumn{1}{c}{Prec} &
  \multicolumn{1}{c}{Recall} &
  \multicolumn{1}{c}{F1}  \\ \hline 

PGDM~\cite{PGDM} & ICME18 & 82.97 & 78.08 & 86.86 & 84.68 & 85.76 & 74.95 & 73.08 & 84.36 & 82.84 & 83.29 & 74.31 & 64.57 & 78.86 & 75.90 & 77.35 \\ 	
SSCsoft~\cite{sscr} & ICCV21 & 86.52 & 78.95 & 86.02 & 87.12 & 86.99  & 81.87 & 78.89 & 85.98 & 89.10 & 86.87 & 82.77 & 68.37 & 75.05 & \textbf{87.49} & 80.43 \\ 				
IAA~\cite{IAA} &  PR22 & 85.27 & 78.04 & 86.08 & 85.80 & 85.64  & 81.94 & 80.31 & 88.36 & 88.01 & 87.80 & 81.72 & 68.47 & 79.56 & 82.06 & 80.37 \\
CAS~\cite{cas} &  IJCV22 & 86.40 & 79.93 & 87.03 & 87.33 & 87.18  & 82.86 & 79.64 & 86.81 & 87.79 & 86.40 & \textbf{84.18} & 68.59 & 77.56 & 83.81 & 80.56 \\
VTB~\cite{VTB} & TCSVT22 & 85.31 & 79.60 & 86.76 & 87.17 & 86.71 & \underline{83.72} & 80.89 & 87.88 & \textbf{89.30} & 88.21 & 82.67 & 69.44 & 78.28 & 84.39 & 80.84 \\
DAFL~\cite{dafl} & AAAI22 & 87.07 & 78.88 & 85.78 & 87.03 & 86.40  & 83.54 & 80.13 & 87.01 & \underline{89.19} & 88.09 & \underline{83.72} & 68.18 & 77.41 & 83.39 & 80.29 \\
Label2Label~\cite{l2l} & ECCV22 & - & - & - & - & -  & 82.24 & 79.23 & 86.39 & 88.57 & 87.08 & - & - & - & - & - \\
SOFA~\cite{sofa} & AAAI24 & \underline{87.10} & \underline{81.10} & \underline{87.80} &\underline{88.40} & \underline{87.80}  & 83.40 & \underline{81.10} & \underline{88.40} & 89.00 & \underline{88.30} & 83.40 & \underline{70.00} & \underline{80.00} & 83.00 & \textbf{81.20} \\
\hline
POAR~\cite{POAR} & MM23 & 83.10 & - & - & - & 84.40  & - & - & - & - & - & - & - & - & - & - \\
POAR\textsuperscript{*}~\cite{POAR} & MM23 & 83.24 & 78.56 & 86.43  & 85.01 & 85.43  & 81.25 & 79.26 & 85.37 & 85.64 & 85.12 & 81.54 & 68.26 & 78.21 & 82.38 & 80.04 \\
\hline
FOCUS (Ours) & - & \textbf{88.04} & \textbf{81.96} & \textbf{88.56} & \textbf{89.07} & \textbf{88.54}   & \textbf{83.90} & \textbf{81.23} & \textbf{89.29} & 88.97 & \textbf{88.41} & 83.45 & \textbf{70.14} & \textbf{80.10} & \underline{85.18} & \underline{80.91}  \\
\hline \toprule [0.5 pt] 
\end{tabular}}
\vspace{-1.5em}
\end{table*}
\section{Experiments}

\subsection{Experimental Setting}

\subsubsection{Datasets and Evaluation Protocols}
We evaluate our method on three publicly available pedestrian attribute recognition datasets, including PETA\cite{peta}, PA100K\cite{attn4}, and RAPv1\cite{rapv1}. The introduction to these datasets is as follows:

\begin{itemize}
    \item The PETA\cite{peta} dataset contains 9500 pedestrian images which 7600 images for training and 1900 images for testing. Following the official protocol\cite{peta}, 35 binary attributes are adopted to evaluate the performance.
    \item The PA100K\cite{attn4} dataset contains 100,000 pedestrian images and is split into training, validation, and test sets with a ratio of 8:1:1. Each image is annotated with 26 commonly used attributes. 
    \item The RAPv1\cite{rapv1} dataset contains 41585 pedestrian images which 33268 images for training and 8317 images for testing. Following the official protocol\cite{rapv1}, 51 binary attributes are adopted to evaluate the performance. 
\end{itemize} 

We adopt label-based metric Mean Accuracy (mA), which calculates the classification accuracy for each attribute, respectively, and instance-based metrics (Accuracy, Precision, Recall, and F1 score) for evaluation in a closed-set scenario. To evaluate the attribute recognition performance in the open-domain scenarios. Following POAR\cite{POAR}, we treat the PAR task as an image-to-text retrieval task, and adopt the Recall@K based on image-to-text K-nearest neighbor retrieval.

\subsubsection{Implementation Details} We adopt the visual encoder $\mathcal{V}(\cdot)$ and text encoder $\mathcal{T}(\cdot)$ from CLIP as our backbone. Specifically, the visual encoder is based on the ViT-B/16, the output dimension of the text encoder is 512. The number of heads in cross-attention is 8. The number of global and local Mix Tokens is 8 and 4, respectively. We set m and r to 4. Most of the settings follow POAR\cite{POAR}, including the warmup learning rate, random horizontal flip, and random erasing. Note that the text encoder is frozen during the whole process of training.

\begin{table}[t]
	\centering
	\caption{Comparison with existing methods on PETA, PA100K and RAPv1 datasets in open-domain scenarios. The first highest scores are represented by \textbf{bold} font.}
         \resizebox{0.95\columnwidth}{!}{%
	\begin{tabular}{c|c|cc|cc|cc}
		\toprule
		\multirow{3}[4]{*}{Method} & \multirow{3}[4]{*}{Source Domain} & \multicolumn{6}{c}{Target Domain} \\
		\cmidrule{3-8}          &       & \multicolumn{2}{c|}{PETA} & \multicolumn{2}{c|}{PA100K} & \multicolumn{2}{c}{RAPv1} \\
		&       & R@1   & R@2   & R@1   & R@2   & R@1   & R@2 \\
		\midrule
		CLIP\cite{clip} & –     & 50.2  & 75.7  & 43.4  & 65.9  & 33.6  & 56.5 \\
		VTB\cite{VTB}  & PA100K & 31.4  & 62.2  & 26.9 & 62.2 & 24.2  & 50.7 \\
		\midrule
        POAR\cite{POAR}  & \multirow{2}{*}{PA100K} & 42.3  & 76.2  & 83.3 & 92.6 & \textbf{39.4}  & \textbf{63.6} \\
        FOCUS (Ours) &       & \textbf{51.2}  & \textbf{77.8}  &\textbf{83.7}   & \textbf{95.5}  & 38.7  & 62.9\\
		\midrule
		POAR\cite{POAR}  & \multirow{2}{*}{PETA} & 87.6 & 96.0 & 45.1  & 73.5  & \textbf{42.2} & \textbf{68.6} \\
        FOCUS (Ours) &       & \textbf{88.5}  & \textbf{96.3}     & \textbf{46.3}  & \textbf{74.2} & 41.4 & 67.8 \\
		\midrule
		POAR\cite{POAR}  & \multirow{2}{*}{RAPv1} & 48.8  & 75.0  & 45.1  & 73.1 & 80.6 & 94.4 \\
        FOCUS (Ours) &       & \textbf{50.1}  & \textbf{76.1}   &\textbf{45.7}   & \textbf{73.1}   & \textbf{80.8}  &\textbf{95.3}  \\
		\bottomrule
	\end{tabular}}%
	\label{tab:zero-shot1}%
\end{table}

\subsection{Comparison with State-of-the-art Methods}
We compare our method with the state-of-the-art methods in Table \ref{tab:compare}, typically evaluates performance in a closed-set scenario. We also show the results of the image-to-text retrieval in Table \ref{tab:zero-shot1} to evaluate the performance in open-domain scenarios.

\subsubsection{Closed-Set Scenario}
From Table \ref{tab:compare}, FOCUS achieves state-of-the-art performance on the PETA dataset. Specifically, FOCUS achieves 0.94\%, 0.86\%, 0.76\%, 0.67\%, and 0.74\% performance improvements in mA, Acc, Prec, Recall, and F1, respectively. On the larger-scale dataset, PA100K, FOCUS also obtains the best performance, which improves the mA, Acc, Prec, and F1 by 0.18\%, 0.13\%, 0.89\%, and 0.11\%, respectively. This demonstrates that FOCUS can learn more fine-grained feature representations and achieve a better utilization of larger-scale data. On the RAPv1 dataset, FOCUS achieves comparable performance without employing any external spatial estimation modules. Compared with POAR\cite{POAR}, which is a CLIP-based method and also focuses on fine-grained feature extraction, FOCUS achieves much better performance on three datasets. We owe this to our proposed attribute-guided approach for extracting more precise attribute-level features.

\subsubsection{Open-Domain Scenarios}
As illustrated in Table \ref{tab:zero-shot1}, FOCUS obtains the best image-to-text retrieval performance on three datasets when trained and evaluated on the same dataset. In open-domain scenarios, we only utilize the average of learnable prompts of seen attributes within the same region as prompts for unseen attributes, for simplicity. As we observe that, FOCUS achieves superior results in image-to-text retrieval on the PA100K and PETA datasets, improving Recall@1 by 8.9\% and 1.2\%, respectively, with slightly lower performance on RAPv1. When trained on the RAPv1 dataset and evaluated on the PETA and PA100K datasets, FOCUS also outperforms existing methods by 1.3\% and 0.6\% in Recall@1, respectively. This demonstrates that FOCUS effectively aligns textual attribute features with visual attribute-level features guided by attributes, even for attributes unseen during training. 

\subsection{Ablation Study}
We conduct comprehensive ablation studies on the PA100K dataset to analyze the effectiveness of each component in Table \ref{tab:abla}. RLP represents learnable prompts with regional information. MGMT represents the Multi-Granularity Mix Tokens module. $\text{AVFE}^-$ represents the Attribute-guided Visual Feature Extraction module without Region-Aware Contrastive Learning (RACL) loss. We observe that each component provides an improvement in performance.
\subsubsection{The effectiveness of MGMT and $\text{AVFE}^-$} To evaluate the effectiveness of the MGMT module, we use only the averaged features of multiple Mix Tokens as the output for image representation. We can observe that MGMT improves the mA by 0.54\%, demonstrating that MGMT effectively captures multi-granularity information, contributing to better feature representation. Furthermore, by leveraging the $\text{AVFE}^-$ module to adaptively extract attribute-relevant information, we achieve a further improvement of 1.34\% in mA without applying any constraints.

\subsubsection{The effectiveness of RACL} As shown in the last row of Table \ref{tab:abla}, with the constraint of RACL, FOCUS achieves improvements of 1.04\%, 0.30\%, 0.72\%, 0.38\%, and 0.22\% in mA, Acc, Prec, Recall, and F1, respectively. This demonstrates that RACL enables textual attributes to better focus on the correct Mix Tokens, effectively capturing attribute-relevant information.

\begin{table}[tbp]
	\centering
	\caption{Ablation studies on the effectiveness of each component on PA100K dataset. }
        \resizebox{0.95\columnwidth}{!}{%
	\begin{tabular}{c@{\hskip 3pt}c@{\hskip 3pt}c@{\hskip 3pt}c|ccccc}
		\toprule
		RLP & MGMT    & $\text{AVFE}^-$   & RACL   & mA    & Acc    & Prec  & Recall & F1 \\
		\midrule
        - & - & - & - & 77.84 & 71.77 & 80.70 & 85.73 & 82.25 \\
		$\checkmark$     &       &       &       & 80.98  & 78.38  & 85.49  & 86.20 & 85.49 \\
        $\checkmark$     &  $\checkmark$     &       &       & 81.52  & 79.57  & 87.07  & 86.84 & 86.60\\
	$\checkmark$	&$\checkmark$    & $\checkmark$   &       & 82.86  & 80.93  & 88.57  & 88.59 & 88.19 \\
	$\checkmark$	& $\checkmark$    & $\checkmark$     & $\checkmark$    & \textbf{83.90} & \textbf{81.23} & \textbf{89.29} & \textbf{88.97} & \textbf{88.41}\\
		\bottomrule
	\end{tabular}}%
	\label{tab:abla}%
    \vspace{-1.0em}
\end{table}%

\begin{figure}[tbp]
\centering
\includegraphics[width=0.45\textwidth]{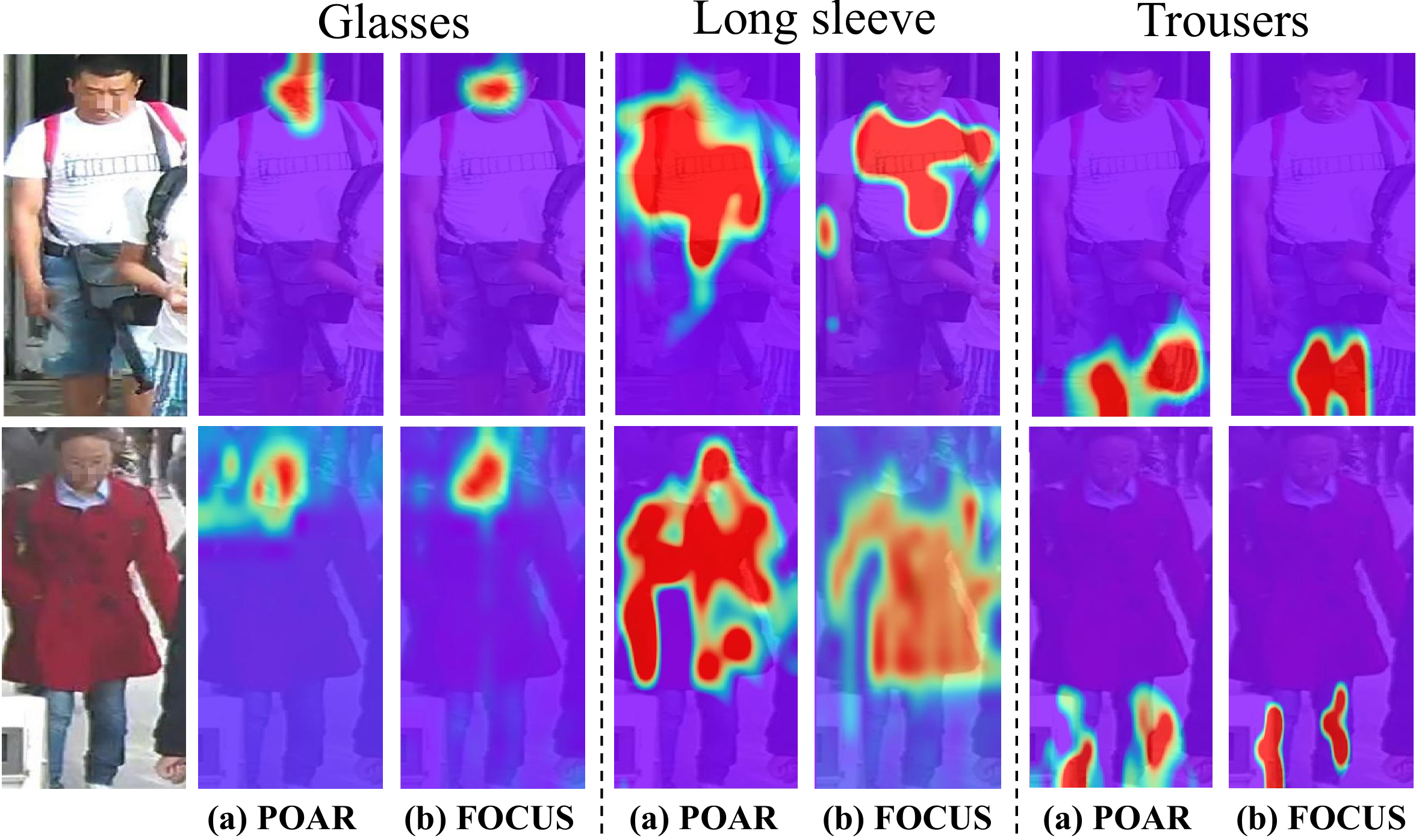} 
\caption{The visualization of attention map for different attributes on PA100K dataset. (a) POAR\cite{POAR}, (b) FOCUS. We can observe that FOCUS can extract attribute-level information and disregard noise guided by attribute.}
\label{attn_vis}
\end{figure}

\subsubsection{The visualization of FOCUS}
Lastly, we visualize the attention maps for different attributes in Fig. \ref{attn_vis}. Compared with POAR\cite{POAR}, which relies on region-level features for attribute recognition, FOCUS leverages attribute guidance to extract attribute-level features. As we can observe, when predicting the attribute of `Long Sleeve', POAR focuses on both the head and upper-body regions, whereas FOCUS precisely separates the attribute of clothes from these regions, capturing more precise visual attribute-level features. Additionally, for attribute of `Trousers', FOCUS effectively disregards occlusions and noise on the right of image, highlighting its strong robustness in complex environments.

\section{Conclusion}
In this paper, we propose FOCUS, a novel framework for pedestrian attribute recognition that is designed to adaptively extract attribute-level feature for each attribute individually, regardless of whether the attributes are seen during training. By leveraging the Multi-Granularity Mix Tokens (MGMT) to capture diverse features and the Attribute-guided Visual Feature Extraction (AVFE) module to extract attribute-related information, FOCUS extracts more precise and adaptive attribute-level features. Furthermore, the Region-Aware Contrastive Learning (RACL) loss refines attribute-level features by focusing on the correct Mix Tokens, significantly enhancing the effectiveness and generalization of attribute-level features in complex scenes. We hope FOCUS can facilitate future work such as cross-modal feature alignment and complex scene understanding in the domain of intelligent transportation.

\bibliographystyle{IEEEbib}
\bibliography{focus}

\end{document}